\pdfoutput=1

\documentclass[11pt]{article}

\usepackage[]{acl}

\usepackage[subtle]{savetrees}
\usepackage{times}
\usepackage{latexsym}
\usepackage{listings}
\usepackage[T1]{fontenc}

\usepackage[utf8]{inputenc}

\usepackage{microtype}
\makeatletter
\MT@redefine@patch{eqnum}{%
  \MT@patch@patch\tagform@{(}{\leftprotrusion{(}}%
  \MT@patch@patch\tagform@{)}{\rightprotrusion{)}}%
  \MT@ifdefined@n@TF{eqref }
    {\MT@exp@cs\MT@patch@patch{eqref }}{\MT@patch@patch\eqref}
       {\tagform@}{\@nameuse{MT@patch@saved@\string\tagform@}}%
}{}
\makeatother

\usepackage{url}
\usepackage{times}
\usepackage{latexsym}

\usepackage{multicol}
\usepackage{multirow}
\usepackage{graphicx}
\usepackage{verbatim}
\usepackage{makecell}
\usepackage{nicefrac, xfrac}
\usepackage{braket}
\usepackage{hhline}
\usepackage{amsmath,bm,amssymb}
\usepackage[linesnumbered,ruled]{algorithm2e}
\usepackage{xcolor}
\definecolor{Green}{rgb}{0.0, 0.5, 0.0}
\definecolor{Amethyst}{rgb}{0.6, 0.4, 0.8}
\title{From Sparse to Dense:\\GPT-4 Summarization with Chain of Density Prompting}

\author{
  Griffin Adams$^{\spadesuit,\clubsuit}$ \\ griffin.adams@columbia.edu \And
  Alexander R. Fabbri$^{\diamondsuit}$ \\ afabbri@salesforce.com \\ \AND Faisal Ladhak $^{\spadesuit}$ \\ faisal@cs.columbia.edu \And Eric Lehman$^{\heartsuit}$ \\ lehmer16@mit.edu \\\And No\'emie Elhadad$^{\spadesuit,\clubsuit}$ \\ noemie.elhadad@columbia.edu \AND
 Salesforce AI$^{\diamondsuit}$ \quad MIT$^{\heartsuit}$ \quad Columbia University: CS$^{\spadesuit}$, Biomedical Informatics$^{\clubsuit}$
}

\begin{document}
\maketitle
\begin{abstract}

Selecting the ``right'' amount of information to include in a summary is a difficult task. A good summary should be detailed and entity-centric without being overly dense and hard to follow. To better understand this tradeoff, we solicit increasingly dense GPT-4 summaries with what we refer to as a ``Chain of Density'' (CoD) prompt. Specifically, GPT-4 generates an initial entity-sparse summary before iteratively incorporating missing salient entities without increasing the length. Summaries generated by CoD are more abstractive, exhibit more fusion, and have less of a lead bias than GPT-4 summaries generated by a vanilla prompt. We conduct a human preference study on 100 CNN DailyMail articles and find that that humans prefer GPT-4 summaries that are more dense than those generated by a vanilla prompt and almost as dense as human written summaries. Qualitative analysis supports the notion that there exists a tradeoff between informativeness and readability. 500 annotated CoD summaries, as well as an extra 5,000 unannotated summaries, are freely available on HuggingFace\footnote{\url{https://huggingface.co/datasets/griffin/chain_of_density}}.

\end{abstract}

\section{Introduction}

Automatic summarization has come a long way in the past few years, largely due to a paradigm shift away from supervised fine-tuning on labeled datasets to zero-shot prompting with Large Language Models (LLMs), such as GPT-4 \citep{gpt4}. Without additional training, careful prompting can enable fine-grained control over summary characteristics, such as length \citep{goyal2022news}, topics \citep{bhaskar-etal-2023-prompted}, and style \citep{pu-demberg-2023-chatgpt}.

An overlooked aspect is the information density of an summary. In theory, as a compression of another text, a summary \emph{should} be denser--containing a higher concentration of information--than the source document. Given the high latency of LLM decoding \citep{kaddour2023challenges}, covering more information in fewer words is a worthy goal, especially for real-time applications. Yet, how dense is an open question. A summary is uninformative if it contains insufficient detail. If it contains too much information, however, it can become difficult to follow without having to increase the overall length. Conveying more information subject to a fixed token budget requires a combination of abstraction, compression, and fusion. There is a limit to how much space can be made for additional information before becoming illegible or even factually incorrect.

\begin{figure}[t]
\centering
\includegraphics[width=\linewidth]{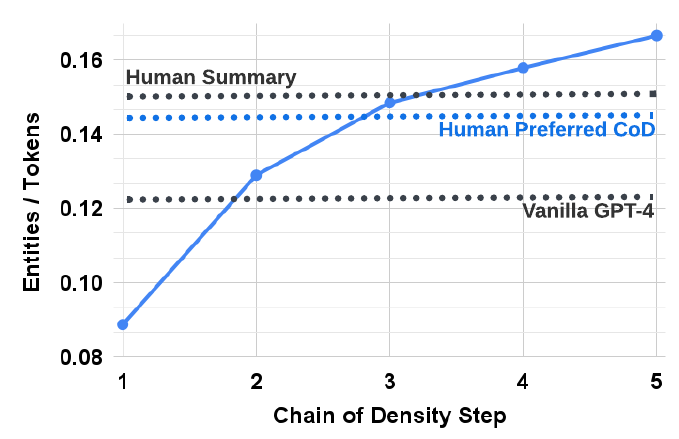}
\caption{Chain of Density (\texttt{\textbf{CoD}}) summaries grow increasingly entity dense, starting off closer to vanilla GPT-4 summaries and eventually surpassing that of human written summaries. Human annotations suggest that a density similar to that of human-written summaries is preferable--striking the right balance between clarity (favors \emph{less} dense) and informativeness (favors \emph{more} dense).}
\label{fig:density}
\end{figure}

\begin{figure*}[t]
\centering
\includegraphics[width=\linewidth]{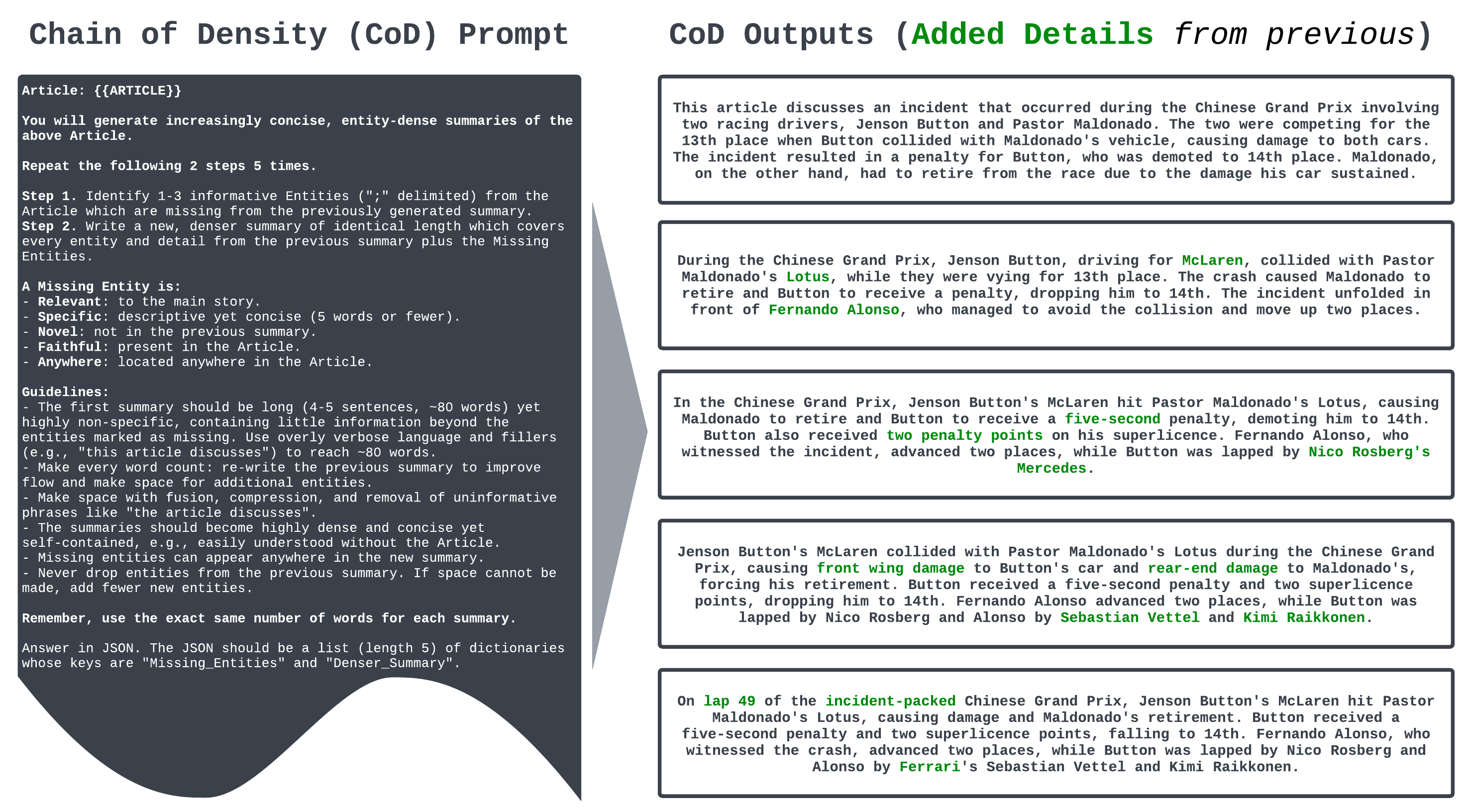}
\caption{Chain of Density (\texttt{\textbf{CoD}}) Prompt and example output. At each step, 1-3 additional details (entities) are added to the previous summary without increasing the length. To make room for new entities, existing content is re-written (e.g., compression, fusion). Half the annotators (2/4) prefer the second to last summary, with the others preferring the final one. }
\label{fig:prompt}
\end{figure*}

In this paper, we seek to identify this limit by soliciting human preferences on a set of increasingly dense summaries produced by GPT-4. Treating entities, and, in particular, the average number of entities per token, as a proxy for density, we generate an initial, entity-sparse summary. Then, we iteratively identify and fuse 1-3 missing entities from the previous summary without increasing the overall length ($5x$ overall). Each summary has a higher ratio of entities to tokens than the previous one. Based on human preference data, we determine that humans prefer summaries that are almost as dense as human-written summaries and more dense than those generated by a vanilla GPT-4 prompt.

\noindent Our primary contributions are to:

\begin{itemize}
    \setlength\itemsep{-0.3em}
    \item Develop a prompt-based iterative method (\texttt{\textbf{CoD}}) for making summaries increasingly entity dense.
    \item Conduct both human and automatic evaluation of increasingly dense summaries on CNN/Dailymail articles to better understand the tradeoff between informativeness (favoring more entities) and clarity (favoring fewer entities).
    \item Open source GPT-4 summaries, annotations, and a set of 5,000 unannotated \texttt{\textbf{CoD}} summaries to be used for evaluation or distillation.
\end{itemize}

\begin{figure*}[t]
\centering
\includegraphics[width=\linewidth]{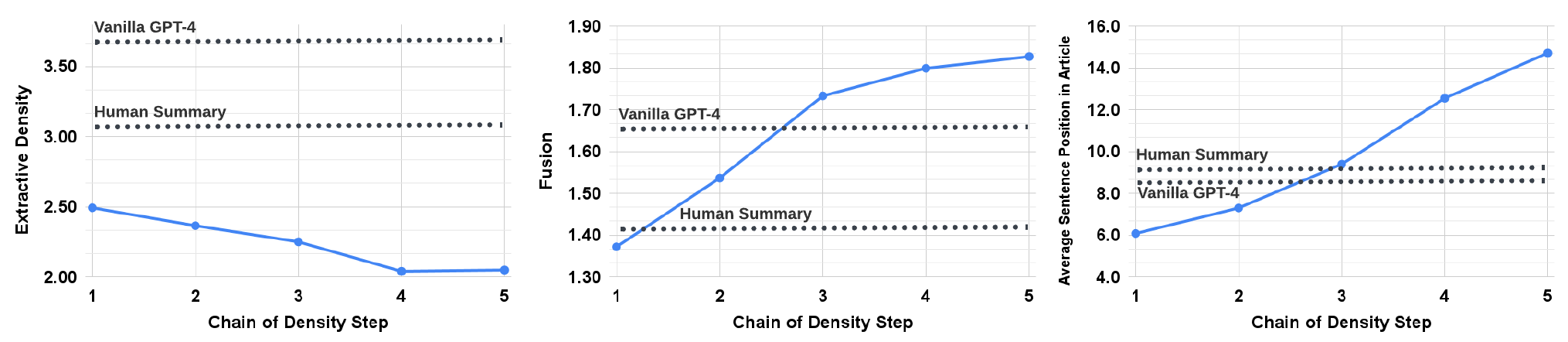}
\caption{\texttt{\textbf{CoD}}-generated summaries grow increasingly abstractive while exhibiting more fusion and less of a lead bias.}
\label{fig:implicit-stats}
\end{figure*}

\section{Chain of Density Prompting} \label{sec:prompt}

\paragraph{Prompt.} Our goal is to generate a set of summaries with GPT-4 with varying levels of information density, while controlling for length, which has proven to be a strong confounder when evaluating summaries \citep{fabbri-etal-2021-summeval, liu-etal-2023-revisiting}. To do this, we formulate a single Chain of Density (\textbf{\texttt{CoD}}) prompt, whereby an initial summary is generated and made increasingly entity dense. Specifically, for a fixed number of turns, a set of unique salient entities from the source text are identified and fused into the previous summary without increasing the length. The first summary is entity-sparse as it focuses on only 1-3 initial entities. To maintain the same length while increasing the number of entities covered, abstraction, fusion, and compression is explicitly encouraged, rather than dropping meaningful content from previous summaries.

Figure \ref{fig:prompt} displays the prompt along with an example output. Rather than be prescriptive about the types of entities, we simply define a Missing Entity as:

\begin{itemize}
    \setlength\itemsep{-0.5em}
    \item \textbf{Relevant}: to the main story.
    \item \textbf{Specific}: descriptive yet concise (5 words or fewer).
    \item \textbf{Novel}: not in the previous summary.
    \item \textbf{Faithful}: present in the Article.
    \item \textbf{Anywhere}: located anywhere in the Article.
\end{itemize}

\paragraph{Data.} We randomly sample 100 articles from the CNN/DailyMail summarization \citep{cnn} test set for which to generate \texttt{\textbf{CoD}} summaries.

\paragraph{Reference Points.}  For frame of reference, we compare \texttt{\textbf{CoD}} summary statistics to human-written bullet-point style reference summaries as well as summaries generated by GPT-4 with a vanilla prompt: ``Write a VERY short summary of the Article. Do not exceed 70 words.'' We set the desired token length to match that of \texttt{\textbf{CoD}} summaries (shown in Table \ref{tab:explicit}).

\section{Statistics} \label{sec:statistics}

Direct statistics (tokens, entities, entity density) are ones directly controlled for by \texttt{\textbf{CoD}}, while Indirect statistics are expected byproducts of densification.

\begin{table}[h]
\centering
\small
\begin{tabular}{c|cc|c}
\textbf{\texttt{CoD} Step} & \textbf{Tokens} & \textbf{Entities} & \textbf{Density (E/T)} \\ \hline
\textbf{1} & 72 & 6.4 & 0.089 \\
\textbf{2} & 67 & 8.7 & 0.129 \\
\textbf{3} & 67 & 9.9 & 0.148 \\
\textbf{4} & 69 & 10.8 & 0.158 \\
\textbf{5} & 72 & 12.1 & 0.167 \\ \hline \hline
\textbf{Human} & 60 & 8.8 & 0.151 \\
\textbf{Vanilla GPT-4} & 70 & 8.5 & 0.122 \\
\end{tabular}
\caption{Explicit statistics for GPT-4 \texttt{\textbf{CoD}} summaries. } \label{tab:explicit}
\end{table}

\paragraph{Direct Statistics.} In Table \ref{tab:explicit}, we compute tokens with NLTK \citep{loper2002nltk}, measure unique entities with Spacy\footnote{\url{https://spacy.io}.}, and compute entity density as the ratio. The \texttt{\textbf{CoD}} prompt largely adheres to a fixed token budget. In fact, the second step leads to an average 5-token (72 to 67) reduction in length as unnecessary words are removed from the initially verbose summary. The entity density rises--starting at $0.089$, initially below Human and Vanilla GPT-4 ($0.151$ and $0.122$)--to $0.167$ after 5 steps of densification.

\paragraph{Indirect Statistics.} \textbf{\emph{Abstractiveness}} should increase with each \texttt{\textbf{CoD}} step because summaries are iteratively re-written to make space for each additional entity. We measure abstractiveness with extractive density: the average squared length of extractive fragments \citep{grusky-etal-2018-newsroom}. Similarly, the level of concept \textbf{\emph{Fusion}} should increase monotonically as entities are added to a fixed-length summary. We proxy fusion as average number of source sentences aligned to each summary sentence. For alignment, we use the relative ROUGE gain method \citep{zhou-etal-2018-neural-document}, which aligns source sentences to a target sentence until the relative ROUGE gain of an additional sentence is no longer positive. We also expect the \textbf{\emph{Content Distribution}}--the position in the Article from which summary content is sourced--to shift. Specifically, we expect that \texttt{\textbf{CoD}} summaries initially exhibit a strong Lead Bias yet gradually start to pull in entities from the middle and end of the article. To measure this, we use our alignments from fusion and measure the average sentence rank of all aligned source sentences. Figure \ref{fig:implicit-stats} confirms these hypotheses: abstractiveness increases with the number of re-writing steps (lower extractive density on the left), the rate of fusion rises (middle figure), and the summaries start to incorporate content from the middle and end of the article (right figure). Interestingly, all \texttt{\textbf{CoD}} summaries are more abstractive than both human written and baseline summaries.

\section{Results} \label{sec:results}

To better understand the tradeoffs present with \texttt{\textbf{CoD}} summaries, we conduct a preference-based human study and a rating-based evaluation with GPT-4.

\begin{table}[h]
\centering
\small
\begin{tabular}{c|ccccc}
\texttt{\textbf{CoD}} & \multicolumn{5}{c}{\textbf{\% Share of First Place Votes}} \\
\textbf{Step} & \multicolumn{4}{c}{\textbf{Individual Annotators}} & \textbf{Aggregate} \\ \hline
\textbf{1} & 3.0 & 2.0 & 13.0 & 17.4 & 8.3 \\
\textbf{2} & 25.0 & \textbf{28.0} & \textbf{43.0} & \textbf{31.4} & \textbf{30.8} \\
\textbf{3} & 22.0 & \textbf{28.0} & 21.0 & 24.4 & 23.0 \\
\textbf{4} & \textbf{29.0} & 25.0 & 13.0 & 26.7 & 22.5 \\
\textbf{5} & 21.0 & 17.0 & 10.0 & 16.3 & 15.5 \\
\hline
\end{tabular}
\caption{Breakdown of first-place votes for \texttt{\textbf{CoD}} summaries by step. Based on aggregate preferences, the modal \texttt{\textbf{CoD}} step is \textbf{2}, median is \textbf{3}, and expected is \textbf{3.06}. } \label{tab:human-preferences}
\end{table}

\begin{table*}[t]
\centering 
\small
\begin{tabular}{cc|ccccc|c}
\textbf{\texttt{CoD} Step} & \textbf{Entity Density} & \textbf{Informative} & \textbf{Quality} & \textbf{Coherence} & \textbf{Attributable} & \textbf{Overall} & \textbf{GPT-4 Eval Average} \\ \hline
\textbf{1} & 0.089 & 4.34 & 4.75 & \textbf{4.96} & 4.96 & 4.41 & 4.69 \\
\textbf{2} & 0.129 & 4.62 & \textbf{4.79} & 4.92 & \textbf{5.00} & 4.58 & \textbf{4.78} \\
\textbf{3} & 0.148 & 4.67 & 4.76 & 4.84 & \textbf{5.00} & 4.57 & 4.77 \\
\textbf{4} & 0.158 & \textbf{4.74} & 4.69 & 4.75 & \textbf{5.00} & \textbf{4.61} & 4.76 \\
\textbf{5} & 0.167 & 4.73 & 4.65 & 4.61 & 4.97 & 4.58 & 4.71 \\
\hline
\end{tabular}
\caption{GPT-4 Likert-scale (1-5) assessments of Chain of Density  (\texttt{\textbf{CoD}}) Summaries by step. } \label{tab:gpt4-likert}
\end{table*}

\begin{figure*}
\centering
\includegraphics[width=\linewidth]{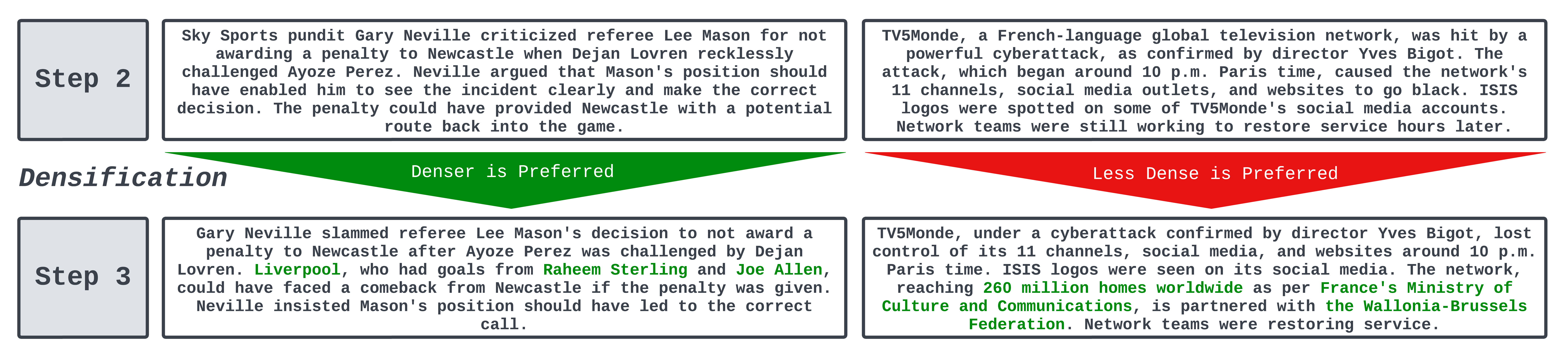}
\caption{An example of a human-preferred densification step (left) and one which is not preferred. For the left, the bottom summary is preferred because the addition of ``Liverpool'' and the goal-scorers is relevant. The second summary makes room with sensible compressions, such as synthesizing ``a potential route back into the game'' into ``a comeback''.  For the right, the addition of more details on ``TVMonde'' does not make up for the presence of an awkward fusion of entities (``cyberattack'', and ``Yves Bigot''), which was a direct result of having to tighten the previous summary. }
\label{fig:qualitative}
\end{figure*}

\paragraph{Human Preferences.} We conduct a human evaluation to assess the impact of densification on human assessments of overall quality. Specifically, the first four authors of the paper were presented with randomly shuffled \texttt{\textbf{CoD}} summaries, along with the articles, for the same 100 articles (5 steps * 100 = 500 total summaries). Based on the definition of a ``good summary" from \citet{rlhf} (Table 6 from their paper), each annotator indicated their top preferred summary. Table \ref{tab:human-preferences} reports the breakdown of first place votes by \texttt{\textbf{CoD}} step across annotators--as well as aggregated across annotators. First, we report a low Fleiss' kappa \citep{fleiss1971measuring} of $0.112$, which points to the subtle differences between summaries and the subjective nature of the task. Recent work has similarly noted low instance-level agreement when judging GPT-based summaries \cite{goyal2022news}.

Yet, at the system level, some trends start to emerge. For 3 of the 4 annotators, \texttt{\textbf{CoD}} step 1 received the largest share of first-place votes across the 100 examples (28, 43, and 31.4\%, respectively). Yet, in aggregate, 61\% of first placed summaries ($23.0 + 22.5 + 15.5$) involved $\geq 3$ densification steps. The median preferred \texttt{\textbf{CoD}} step is in the middle ($3$), and the expected step is $3.06$.


Based on the average density of Step 3 summaries, we can roughly infer a preferred entity density of $\sim 0.15$ across the \texttt{\textbf{CoD}} candidates. From Table \ref{tab:explicit}, we can see that this density aligns with human-written summaries ($0.151$), yet is noticeable higher than summaries produced with a vanilla GPT-4 prompt ($0.122$).

\paragraph{Automatic Metrics.} As an evaluator, GPT-4 has been shown to adequately correlate to human judgments \citep{fu2023gptscore, liu2023gpteval}, even potentially outperforming crowd-sourced workers on some annotation tasks \citep{gilardi2023chatgpt}. As a complement to our human evaluation (below), we prompt GPT-4 to rate \texttt{\textbf{CoD}} summaries (1-5) along 5 dimensions: \textbf{Informative}, \textbf{Quality}, \textbf{Coherence}, \textbf{Attributable}, and \textbf{Overall}. The definitions of \textbf{Informative}, \textbf{Quality}, and \textbf{Attributable} come from \citet{aharoni-etal-2023-multilingual}, while \textbf{Coherence} comes from \citet{fabbri-etal-2021-summeval}\footnote{\textbf{Quality} and \textbf{Coherence} are article-independent metrics.}. \textbf{Overall} aims to capture the qualities jointly. Please see Appendix \ref{app:metrics} for the prompts used to solicit scores for each dimension. Table \ref{tab:gpt4-likert} suggests that densification is correlated with informativeness, yet there is a limit, with the score peaking at Step 4 ($4.74$). Article-free dimensions: \textbf{Quality} and \textbf{Coherence}, decline sooner (after 2 and 1 steps, respectively). All summaries are deemed \textbf{Attributable} to the source article. The \textbf{Overall} scores skew toward denser and more informative summaries, with \textbf{Step 4} having the highest score. On average across dimensions, the first and last \texttt{\textbf{CoD}} steps are \emph{least} favored, while the middle three are close ($4.78$, $4.77$, and $4.76$, respectively).

In Appendix \ref{app:metrics}, we report highest summary-level correlations of the \textbf{Overall} metric to human judgments (0.31 Pearson correlation), yet note low correlations overall--a phenomenon observed by \citet{deutsch-etal-2022-examining} when summaries are of similar quality.

\paragraph{Qualitative Analysis.} There exists a clear trade-off between coherence / readability of summaries and informativeness.  To illustrate, in Figure \ref{fig:qualitative}, we present two \texttt{\textbf{CoD}} steps: one for which the summary is improved with more detail, and one for which the summary is harmed. On average, intermediate \texttt{\textbf{CoD}} summaries best achieved this balance, yet we leave it to future work to precisely define and quantify this tradeoff.

\section{Related Work} \label{sec:related}

\textbf{GPT Summarization.} \citet{goyal2022news} benchmarked GPT-3 on news article summarization and found that humans preferred GPT-3 summaries over previous supervised baselines, which was not reflective of existing reference-based and reference-free metrics. \citet{zhang2023benchmarking} find that zeroshot GPT-3 summaries perform on par with humans by soliciting high-quality summaries from freelance writers. \textbf{Entity-Based Summarization.} \citet{narayan-etal-2021-planning} proposed generating entity chains as a planning step for supervised fine-tuning of summarization models, in contrast to keywords \citep{li2020keywords, dou-etal-2021-gsum} or purely extractive units \citep{dou-etal-2021-gsum, adams-etal-2023-generating}. Entities have also been incorporated for summarization as a form of control \citep{liu-chen-2021-controllable, he-etal-2022-ctrlsum, maddela-etal-2022-entsum}, to improve faithfulness \citep{nan-etal-2021-entity, adams-etal-2022-learning}, and as a unit for evaluation \citep{cao-etal-2022-hallucinated, adams2023meta}.

\section{Conclusion} \label{sec:conclusion}

We study the impact of summary densification on human preferences of overall quality. We find that a degree of densification is preferred, yet, when summaries contain too many entities per token, it is very difficult maintain readability and coherence. We open-source annotated test set as well as a larger un-annotated training set for further research into the topic of fixed-length, variable density summarization.

\section{Limitations} \label{sec:limitations}

We only analyze \textbf{\texttt{CoD}} for a single domain, news summarization. Annotations did not show high summary-level agreement yet did start to show system-level trends, which is in line with previous work on LLM-based evaluation \citep{goyal2022news}. Finally, GPT-4 is a closed source model so we cannot share model weights. We do, however, publish all evaluation data, annotations, as well as $5,000$ un-annotated \textbf{\texttt{CoD}} to be used for downstream uses cases, e.g., density distillation into an open-sourced model such as LLAMA-2 \citep{touvron2023llama}.

\bibliography{anthology,custom}

\begin{thebibliography}{31}
\expandafter\ifx\csname natexlab\endcsname\relax\def\natexlab#1{#1}\fi

\bibitem[{Adams et~al.(2023{\natexlab{a}})Adams, Fabbri, Ladhak, Elhadad, and
  McKeown}]{adams-etal-2023-generating}
Griffin Adams, Alex Fabbri, Faisal Ladhak, No{\'e}mie Elhadad, and Kathleen
  McKeown. 2023{\natexlab{a}}.
\newblock \href {https://aclanthology.org/2023.acl-long.151} {Generating {EDU}
  extracts for plan-guided summary re-ranking}.
\newblock In \emph{Proceedings of the 61st Annual Meeting of the Association
  for Computational Linguistics (Volume 1: Long Papers)}, pages 2680--2697,
  Toronto, Canada. Association for Computational Linguistics.

\bibitem[{Adams et~al.(2022)Adams, Shing, Sun, Winestock, McKeown, and
  Elhadad}]{adams-etal-2022-learning}
Griffin Adams, Han-Chin Shing, Qing Sun, Christopher Winestock, Kathleen
  McKeown, and No{\'e}mie Elhadad. 2022.
\newblock \href {https://aclanthology.org/2022.findings-emnlp.296} {Learning to
  revise references for faithful summarization}.
\newblock In \emph{Findings of the Association for Computational Linguistics:
  EMNLP 2022}, pages 4009--4027, Abu Dhabi, United Arab Emirates. Association
  for Computational Linguistics.

\bibitem[{Adams et~al.(2023{\natexlab{b}})Adams, Zucker, and
  Elhadad}]{adams2023meta}
Griffin Adams, Jason Zucker, and No{\'e}mie Elhadad. 2023{\natexlab{b}}.
\newblock A meta-evaluation of faithfulness metrics for long-form
  hospital-course summarization.
\newblock \emph{arXiv preprint arXiv:2303.03948}.

\bibitem[{Aharoni et~al.(2023)Aharoni, Narayan, Maynez, Herzig, Clark, and
  Lapata}]{aharoni-etal-2023-multilingual}
Roee Aharoni, Shashi Narayan, Joshua Maynez, Jonathan Herzig, Elizabeth Clark,
  and Mirella Lapata. 2023.
\newblock \href {https://aclanthology.org/2023.findings-acl.220} {Multilingual
  summarization with factual consistency evaluation}.
\newblock In \emph{Findings of the Association for Computational Linguistics:
  ACL 2023}, pages 3562--3591, Toronto, Canada. Association for Computational
  Linguistics.

\bibitem[{Bhaskar et~al.(2023)Bhaskar, Fabbri, and
  Durrett}]{bhaskar-etal-2023-prompted}
Adithya Bhaskar, Alex Fabbri, and Greg Durrett. 2023.
\newblock \href {https://aclanthology.org/2023.findings-acl.591} {Prompted
  opinion summarization with {GPT}-3.5}.
\newblock In \emph{Findings of the Association for Computational Linguistics:
  ACL 2023}, pages 9282--9300, Toronto, Canada. Association for Computational
  Linguistics.

\bibitem[{Cao et~al.(2022)Cao, Dong, and Cheung}]{cao-etal-2022-hallucinated}
Meng Cao, Yue Dong, and Jackie Cheung. 2022.
\newblock \href {https://doi.org/10.18653/v1/2022.acl-long.236} {Hallucinated
  but factual! inspecting the factuality of hallucinations in abstractive
  summarization}.
\newblock In \emph{Proceedings of the 60th Annual Meeting of the Association
  for Computational Linguistics (Volume 1: Long Papers)}, pages 3340--3354,
  Dublin, Ireland. Association for Computational Linguistics.

\bibitem[{Deutsch et~al.(2022)Deutsch, Dror, and
  Roth}]{deutsch-etal-2022-examining}
Daniel Deutsch, Rotem Dror, and Dan Roth. 2022.
\newblock \href {https://doi.org/10.18653/v1/2022.naacl-main.442} {Re-examining
  system-level correlations of automatic summarization evaluation metrics}.
\newblock In \emph{Proceedings of the 2022 Conference of the North American
  Chapter of the Association for Computational Linguistics: Human Language
  Technologies}, pages 6038--6052, Seattle, United States. Association for
  Computational Linguistics.

\bibitem[{Dou et~al.(2021)Dou, Liu, Hayashi, Jiang, and
  Neubig}]{dou-etal-2021-gsum}
Zi-Yi Dou, Pengfei Liu, Hiroaki Hayashi, Zhengbao Jiang, and Graham Neubig.
  2021.
\newblock \href {https://doi.org/10.18653/v1/2021.naacl-main.384} {{GS}um: A
  general framework for guided neural abstractive summarization}.
\newblock In \emph{Proceedings of the 2021 Conference of the North American
  Chapter of the Association for Computational Linguistics: Human Language
  Technologies}, pages 4830--4842, Online. Association for Computational
  Linguistics.

\bibitem[{Fabbri et~al.(2021)Fabbri, Kry{\'s}ci{\'n}ski, McCann, Xiong, Socher,
  and Radev}]{fabbri-etal-2021-summeval}
Alexander~R. Fabbri, Wojciech Kry{\'s}ci{\'n}ski, Bryan McCann, Caiming Xiong,
  Richard Socher, and Dragomir Radev. 2021.
\newblock \href {https://doi.org/10.1162/tacl_a_00373} {{S}umm{E}val:
  Re-evaluating summarization evaluation}.
\newblock \emph{Transactions of the Association for Computational Linguistics},
  9:391--409.

\bibitem[{Fleiss(1971)}]{fleiss1971measuring}
Joseph~L Fleiss. 1971.
\newblock Measuring nominal scale agreement among many raters.
\newblock \emph{Psychological bulletin}, 76(5):378.

\bibitem[{Fu et~al.(2023)Fu, Ng, Jiang, and Liu}]{fu2023gptscore}
Jinlan Fu, See-Kiong Ng, Zhengbao Jiang, and Pengfei Liu. 2023.
\newblock Gptscore: Evaluate as you desire.
\newblock \emph{arXiv preprint arXiv:2302.04166}.

\bibitem[{Gilardi et~al.(2023)Gilardi, Alizadeh, and
  Kubli}]{gilardi2023chatgpt}
Fabrizio Gilardi, Meysam Alizadeh, and Ma{\"e}l Kubli. 2023.
\newblock Chatgpt outperforms crowd-workers for text-annotation tasks.
\newblock \emph{arXiv preprint arXiv:2303.15056}.

\bibitem[{Goyal et~al.(2022)Goyal, Li, and Durrett}]{goyal2022news}
Tanya Goyal, Junyi~Jessy Li, and Greg Durrett. 2022.
\newblock News summarization and evaluation in the era of gpt-3.
\newblock \emph{arXiv preprint arXiv:2209.12356}.

\bibitem[{Grusky et~al.(2018)Grusky, Naaman, and
  Artzi}]{grusky-etal-2018-newsroom}
Max Grusky, Mor Naaman, and Yoav Artzi. 2018.
\newblock \href {https://doi.org/10.18653/v1/N18-1065} {{N}ewsroom: A dataset
  of 1.3 million summaries with diverse extractive strategies}.
\newblock In \emph{Proceedings of the 2018 Conference of the North {A}merican
  Chapter of the Association for Computational Linguistics: Human Language
  Technologies, Volume 1 (Long Papers)}, pages 708--719, New Orleans,
  Louisiana. Association for Computational Linguistics.

\bibitem[{He et~al.(2022)He, Kryscinski, McCann, Rajani, and
  Xiong}]{he-etal-2022-ctrlsum}
Junxian He, Wojciech Kryscinski, Bryan McCann, Nazneen Rajani, and Caiming
  Xiong. 2022.
\newblock \href {https://aclanthology.org/2022.emnlp-main.396} {{CTRL}sum:
  Towards generic controllable text summarization}.
\newblock In \emph{Proceedings of the 2022 Conference on Empirical Methods in
  Natural Language Processing}, pages 5879--5915, Abu Dhabi, United Arab
  Emirates. Association for Computational Linguistics.

\bibitem[{Kaddour et~al.(2023)Kaddour, Harris, Mozes, Bradley, Raileanu, and
  McHardy}]{kaddour2023challenges}
Jean Kaddour, Joshua Harris, Maximilian Mozes, Herbie Bradley, Roberta
  Raileanu, and Robert McHardy. 2023.
\newblock Challenges and applications of large language models.
\newblock \emph{arXiv preprint arXiv:2307.10169}.

\bibitem[{Li et~al.(2020)Li, Zhu, Zhang, Zong, and He}]{li2020keywords}
Haoran Li, Junnan Zhu, Jiajun Zhang, Chengqing Zong, and Xiaodong He. 2020.
\newblock Keywords-guided abstractive sentence summarization.
\newblock In \emph{Proceedings of the AAAI conference on artificial
  intelligence}, volume~34, pages 8196--8203.

\bibitem[{Liu et~al.(2023{\natexlab{a}})Liu, Iter, Xu, Wang, Xu, and
  Zhu}]{liu2023gpteval}
Yang Liu, Dan Iter, Yichong Xu, Shuohang Wang, Ruochen Xu, and Chenguang Zhu.
  2023{\natexlab{a}}.
\newblock Gpteval: Nlg evaluation using gpt-4 with better human alignment.
\newblock \emph{arXiv preprint arXiv:2303.16634}.

\bibitem[{Liu et~al.(2023{\natexlab{b}})Liu, Fabbri, Liu, Zhao, Nan, Han, Han,
  Joty, Wu, Xiong, and Radev}]{liu-etal-2023-revisiting}
Yixin Liu, Alex Fabbri, Pengfei Liu, Yilun Zhao, Linyong Nan, Ruilin Han,
  Simeng Han, Shafiq Joty, Chien-Sheng Wu, Caiming Xiong, and Dragomir Radev.
  2023{\natexlab{b}}.
\newblock \href {https://aclanthology.org/2023.acl-long.228} {Revisiting the
  gold standard: Grounding summarization evaluation with robust human
  evaluation}.
\newblock In \emph{Proceedings of the 61st Annual Meeting of the Association
  for Computational Linguistics (Volume 1: Long Papers)}, pages 4140--4170,
  Toronto, Canada. Association for Computational Linguistics.

\bibitem[{Liu and Chen(2021)}]{liu-chen-2021-controllable}
Zhengyuan Liu and Nancy Chen. 2021.
\newblock \href {https://doi.org/10.18653/v1/2021.emnlp-main.8} {Controllable
  neural dialogue summarization with personal named entity planning}.
\newblock In \emph{Proceedings of the 2021 Conference on Empirical Methods in
  Natural Language Processing}, pages 92--106, Online and Punta Cana, Dominican
  Republic. Association for Computational Linguistics.

\bibitem[{Loper and Bird(2002)}]{loper2002nltk}
Edward Loper and Steven Bird. 2002.
\newblock Nltk: The natural language toolkit.
\newblock \emph{arXiv preprint cs/0205028}.

\bibitem[{Maddela et~al.(2022)Maddela, Kulkarni, and
  Preotiuc-Pietro}]{maddela-etal-2022-entsum}
Mounica Maddela, Mayank Kulkarni, and Daniel Preotiuc-Pietro. 2022.
\newblock \href {https://doi.org/10.18653/v1/2022.acl-long.237} {{E}nt{SUM}: A
  data set for entity-centric extractive summarization}.
\newblock In \emph{Proceedings of the 60th Annual Meeting of the Association
  for Computational Linguistics (Volume 1: Long Papers)}, pages 3355--3366,
  Dublin, Ireland. Association for Computational Linguistics.

\bibitem[{Nallapati et~al.(2016)Nallapati, Zhou, dos Santos, Gul{\c{c}}ehre,
  and Xiang}]{cnn}
Ramesh Nallapati, Bowen Zhou, Cicero dos Santos, {\c{C}}a{\u{g}}lar
  Gul{\c{c}}ehre, and Bing Xiang. 2016.
\newblock \href {https://doi.org/10.18653/v1/K16-1028} {Abstractive text
  summarization using sequence-to-sequence {RNN}s and beyond}.
\newblock In \emph{Proceedings of the 20th {SIGNLL} Conference on Computational
  Natural Language Learning}, pages 280--290, Berlin, Germany. Association for
  Computational Linguistics.

\bibitem[{Nan et~al.(2021)Nan, Nallapati, Wang, Nogueira~dos Santos, Zhu,
  Zhang, McKeown, and Xiang}]{nan-etal-2021-entity}
Feng Nan, Ramesh Nallapati, Zhiguo Wang, Cicero Nogueira~dos Santos, Henghui
  Zhu, Dejiao Zhang, Kathleen McKeown, and Bing Xiang. 2021.
\newblock \href {https://doi.org/10.18653/v1/2021.eacl-main.235} {Entity-level
  factual consistency of abstractive text summarization}.
\newblock In \emph{Proceedings of the 16th Conference of the European Chapter
  of the Association for Computational Linguistics: Main Volume}, pages
  2727--2733, Online. Association for Computational Linguistics.

\bibitem[{Narayan et~al.(2021)Narayan, Zhao, Maynez, Sim{\~o}es, Nikolaev, and
  McDonald}]{narayan-etal-2021-planning}
Shashi Narayan, Yao Zhao, Joshua Maynez, Gon{\c{c}}alo Sim{\~o}es, Vitaly
  Nikolaev, and Ryan McDonald. 2021.
\newblock \href {https://doi.org/10.1162/tacl_a_00438} {Planning with learned
  entity prompts for abstractive summarization}.
\newblock \emph{Transactions of the Association for Computational Linguistics},
  9:1475--1492.

\bibitem[{OpenAI(2023)}]{gpt4}
OpenAI. 2023.
\newblock \href {https://api.semanticscholar.org/CorpusID:257532815} {Gpt-4
  technical report}.
\newblock \emph{ArXiv}, abs/2303.08774.

\bibitem[{Pu and Demberg(2023)}]{pu-demberg-2023-chatgpt}
Dongqi Pu and Vera Demberg. 2023.
\newblock \href {https://aclanthology.org/2023.acl-srw.1} {{C}hat{GPT} vs
  human-authored text: Insights into controllable text summarization and
  sentence style transfer}.
\newblock In \emph{Proceedings of the 61st Annual Meeting of the Association
  for Computational Linguistics (Volume 4: Student Research Workshop)}, pages
  1--18, Toronto, Canada. Association for Computational Linguistics.

\bibitem[{Stiennon et~al.(2020)Stiennon, Ouyang, Wu, Ziegler, Lowe, Voss,
  Radford, Amodei, and Christiano}]{rlhf}
Nisan Stiennon, Long Ouyang, Jeffrey Wu, Daniel Ziegler, Ryan Lowe, Chelsea
  Voss, Alec Radford, Dario Amodei, and Paul~F Christiano. 2020.
\newblock Learning to summarize with human feedback.
\newblock \emph{Advances in Neural Information Processing Systems},
  33:3008--3021.

\bibitem[{Touvron et~al.(2023)Touvron, Martin, Stone, Albert, Almahairi,
  Babaei, Bashlykov, Batra, Bhargava, Bhosale et~al.}]{touvron2023llama}
Hugo Touvron, Louis Martin, Kevin Stone, Peter Albert, Amjad Almahairi, Yasmine
  Babaei, Nikolay Bashlykov, Soumya Batra, Prajjwal Bhargava, Shruti Bhosale,
  et~al. 2023.
\newblock Llama 2: Open foundation and fine-tuned chat models.
\newblock \emph{arXiv preprint arXiv:2307.09288}.

\bibitem[{Zhang et~al.(2023)Zhang, Ladhak, Durmus, Liang, McKeown, and
  Hashimoto}]{zhang2023benchmarking}
Tianyi Zhang, Faisal Ladhak, Esin Durmus, Percy Liang, Kathleen McKeown, and
  Tatsunori~B Hashimoto. 2023.
\newblock Benchmarking large language models for news summarization.
\newblock \emph{arXiv preprint arXiv:2301.13848}.

\bibitem[{Zhou et~al.(2018)Zhou, Yang, Wei, Huang, Zhou, and
  Zhao}]{zhou-etal-2018-neural-document}
Qingyu Zhou, Nan Yang, Furu Wei, Shaohan Huang, Ming Zhou, and Tiejun Zhao.
  2018.
\newblock \href {https://doi.org/10.18653/v1/P18-1061} {Neural document
  summarization by jointly learning to score and select sentences}.
\newblock In \emph{Proceedings of the 56th Annual Meeting of the Association
  for Computational Linguistics (Volume 1: Long Papers)}, pages 654--663,
  Melbourne, Australia. Association for Computational Linguistics.

\end{thebibliography}
\bibliographystyle{acl_natbib}

\appendix

\section{GPT-4 Metrics} \label{app:metrics}

For the GPT-4 Likert-style evaluation, we use the following prompt template.

\begin{verbatim}
Article: {{Article}}

Summary: {{Summary}}

Please rate the summary
(1=worst to 5=best) with 
respect to {{Dimension}}. 

{{Definition}}
\end{verbatim}

Below, we present the definitions provided for each quality metric.

\begin{itemize}
    \item \textbf{Informative}: An informative summary captures the important information in the article and presents it accurately and concisely.
    \item \textbf{Quality}: A high quality summary is comprehensible and understandable.
    \item \textbf{Coherence}: A coherent summary is well-structured and well-organized.
    \item \textbf{Attributable}: Is all the information in the summary fully attributable to the Article?
    \item \textbf{Overall Preference}: A good summary should convey the main ideas in the Article in a concise, logical, and coherent fashion.
\end{itemize}

The \textbf{Quality} and \textbf{Coherence} prompts do not include the Article in the prompt. These definitions were paraphrased from previous summarization annotation efforts: \citep{fabbri-etal-2021-summeval, aharoni-etal-2023-multilingual}.

\paragraph{Meta-Evaluation.}

\begin{table}[h]
\centering 
\small
\begin{tabular}{c|c}
\textbf{Dimension} & \textbf{Correlation} \\ \hline
\textbf{Informative} & 0.215 \\
\textbf{Quality} & 0.120 \\
\textbf{Coherence} & 0.178 \\
\textbf{Attributable} & 0.245 \\
\textbf{Overall} & \textbf{0.311} \\
\hline
\end{tabular}
\caption{Summary-Level Pearson Correlation coefficient between human preferences and GPT-4 Likert ratings. } \label{tab:meta}
\end{table}

To compute the summary-level correlation, we first turned the preference data into a vector representing the number of times that summary received a first-placed vote. Table \ref{tab:meta} demonstrates, unsurprisingly, that a prompt designed to capture overall summary rating has the highest summary-level Pearson correlation to overall preferences ($31$), yet overall correlations are still low.

\end{document}